\documentclass{article}
\usepackage{spconf}
\usepackage{graphicx}
\usepackage{threeparttable} 
\usepackage{subcaption} 
\usepackage{booktabs}

\title{Cloud Optical Thickness Retrievals Using Angle Invariant Attention Based Deep Learning Models}

\name{
Zahid Hassan Tushar 
\hspace{3mm}  Adeleke Ademakinwa 
\hspace{3mm}  Jianwu Wang
\hspace{3mm}  Zhibo Zhang
\hspace{3mm}  Sanjay Purushotham}
\address{University of Maryland, Baltimore County, Maryland, USA.
\thanks{© 2025 IEEE. Personal use of this material is permitted. Permission from IEEE must be obtained for all other uses, in any current or future media, including reprinting/republishing this material for advertising or promotional purposes, creating new collective works, for resale or redistribution to servers or lists, or reuse of any copyrighted component of this work in other works.}}

\begin{document}
\maketitle

\begin{abstract}

Cloud Optical Thickness (COT) is a critical cloud property influencing Earth's climate, weather, and radiation budget. Satellite radiance measurements enable global COT retrieval, but challenges like 3D cloud effects, viewing angles, and atmospheric interference must be addressed to ensure accurate estimation. Traditionally, the Independent Pixel Approximation (IPA) method, which treats individual pixels independently, has been used for COT estimation. However, IPA introduces significant bias due to its simplified assumptions. Recently, deep learning-based models have shown improved performance over IPA but lack robustness, as they are sensitive to variations in radiance intensity, distortions, and cloud shadows. These models also introduce substantial errors in COT estimation under different solar and viewing zenith angles. To address these challenges, we propose a novel angle-invariant, attention-based deep model called Cloud-Attention-Net with Angle Coding (CAAC). Our model leverages attention mechanisms and angle embeddings to account for satellite viewing geometry and 3D radiative transfer effects, enabling more accurate retrieval of COT. Additionally, our multi-angle training strategy ensures angle invariance. Through comprehensive experiments, we demonstrate that CAAC significantly outperforms existing state-of-the-art deep learning models, reducing cloud property retrieval errors by at least a factor of nine.

\end{abstract}


\section{Introduction}
Clouds play a crucial role in regulating the Earth's radiation budget and are characterized by various microphysical properties, including cloud optical thickness (COT), cloud effective radius (CER), and cloud top height (CTH). These properties are essential for understanding climate change and improving weather forecasting. Given the wide range of applications involving clouds, accurate estimation of these properties is of great importance. Reflecting this significance, the latest NASA Decadal Survey has identified cloud observations as a top priority for NASA's missions in the upcoming decade~\cite{national2019thriving}. 

Retrieving cloud properties from radiance observations remains inherently challenging due to the 3D radiative transfer effects~\cite{ademakinwa2023influence}. These effects arise from complex interactions between solar radiation and the three-dimensional structure of clouds. Real clouds exhibit intricate shapes and variable densities, leading to scattering, absorption, and emission of radiation across multiple directions in 3D space. In contrast, satellite radiance observations are fundamentally two-dimensional, where each observed point represents an aggregate of radiance contributions from numerous cloud droplets. Consequently, extracting microphysical properties from these observations involves solving a mathematically intractable 3D inverse problem~\cite{nakajima1990determination}.

Another challenge in retrieving cloud properties from radiance data is the effect of solar zenith angle (SZA) and view zenith angle (VZA). Because satellite viewing geometries can vary, the same cloud may appear differently, exhibiting variations in intensity, distortion, and shadow effects—ultimately influencing the retrieved properties. While prior studies have documented the influence of SZA and VZA on satellite spectral reflectances or radiance observations~\cite{marshak2021effect,breunig2015spectral,loeb1997effect}, none have explicitly examined their impact on the performance of cloud property retrieval algorithms.

Estimation of cloud properties from radiance observations, an important topic in remote sensing, has been explored in several pioneering works~\cite{nakajima1990determination,okamura2017feasibility,nataraja2022segmentation,tushar2024cloudunet}. 
Nakajima et al.~\cite{nakajima1990determination} proposed the physics-based IPA retrieval method, which employs a one-dimensional inversion technique for retrieving cloud optical thickness (COT), however, it suffers from significant biases and errors in the retrieval process. 
Recent advances in machine learning and deep learning~\cite{okamura2017feasibility, tushar2024cloudunet, li2023transfer, zhao2023cloud, wang2022retrieval, wang2022cloud, yang2022machine} have demonstrated that data-driven techniques can reduce the gap between retrieved and true cloud properties. However, these methods are designed to retrieve COT from single-angle radiance observations, limiting their applicability to real-world scenarios.
In natural environments, the sun’s position, which determines the solar zenith angle (SZA), and the satellite’s position, which dictates the view zenith angle (VZA), can vary significantly. Therefore, an effective retrieval method must be robust to these angular variations while accurately estimating cloud properties. A straightforward approach  involves training separate models for each angular configuration; however, this would result in significant memory and computational costs.

To address these challenges, we propose a novel model, Cloud-Attention-Net with Angle Coding (CAAC), which is capable of handling diverse angular configurations encountered in real-world scenarios while maintaining both accuracy and computational efficiency. To the best of our knowledge, this is the first deep learning-based model for COT retrieval that effectively mitigates the effects of SZAs and VZAs. Our CAAC model is lightweight and efficiently incorporates angle information through multi-angle training (i.e., training with radiance collected at multiple solar and view zenith angles), allowing it to account for variations in solar and view zenith angles.
Extensive experiments on simulated real-world cloud data demonstrate that our CAAC model outperforms both the IPA method and other deep learning-based retrieval techniques for COT estimation under varying SZAs, VZAs, and multi-angle scenarios—all while incurring lower computational costs.

\section{Data and Problem Formulation}
\begin{figure*}[t]
    \centering
    \begin{subfigure}[t]{0.48\linewidth}
        \centering
        \includegraphics[width=0.99\linewidth]{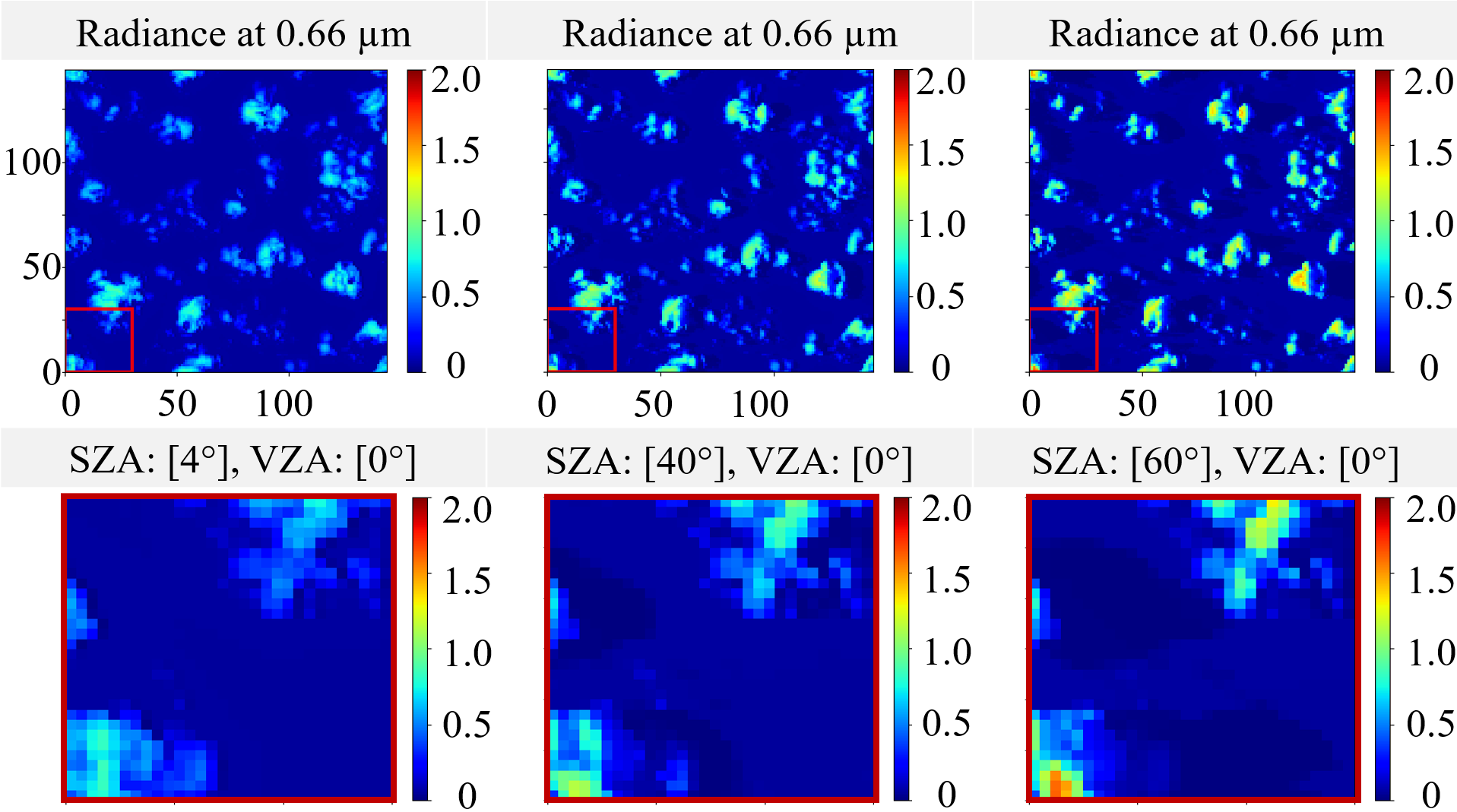}
        \caption{Effects of Solar Zenith Angle (SZA) on satellite radiance observations. 
        Top row: Radiance observations for the \textit{same cloud field} at different SZAs 
        $[4^\circ,40^\circ,60^\circ]$ and nadir VZA $[0^\circ]$. 
        Bottom row: \textit{Highlighted regions} show shadowing and illumination (3D radiative effects) at large SZAs.}
        \label{fig:sza_effects}
    \end{subfigure}
    \hfill
    \begin{subfigure}[t]{0.48\linewidth}
        \centering
        \includegraphics[width=0.99\linewidth]{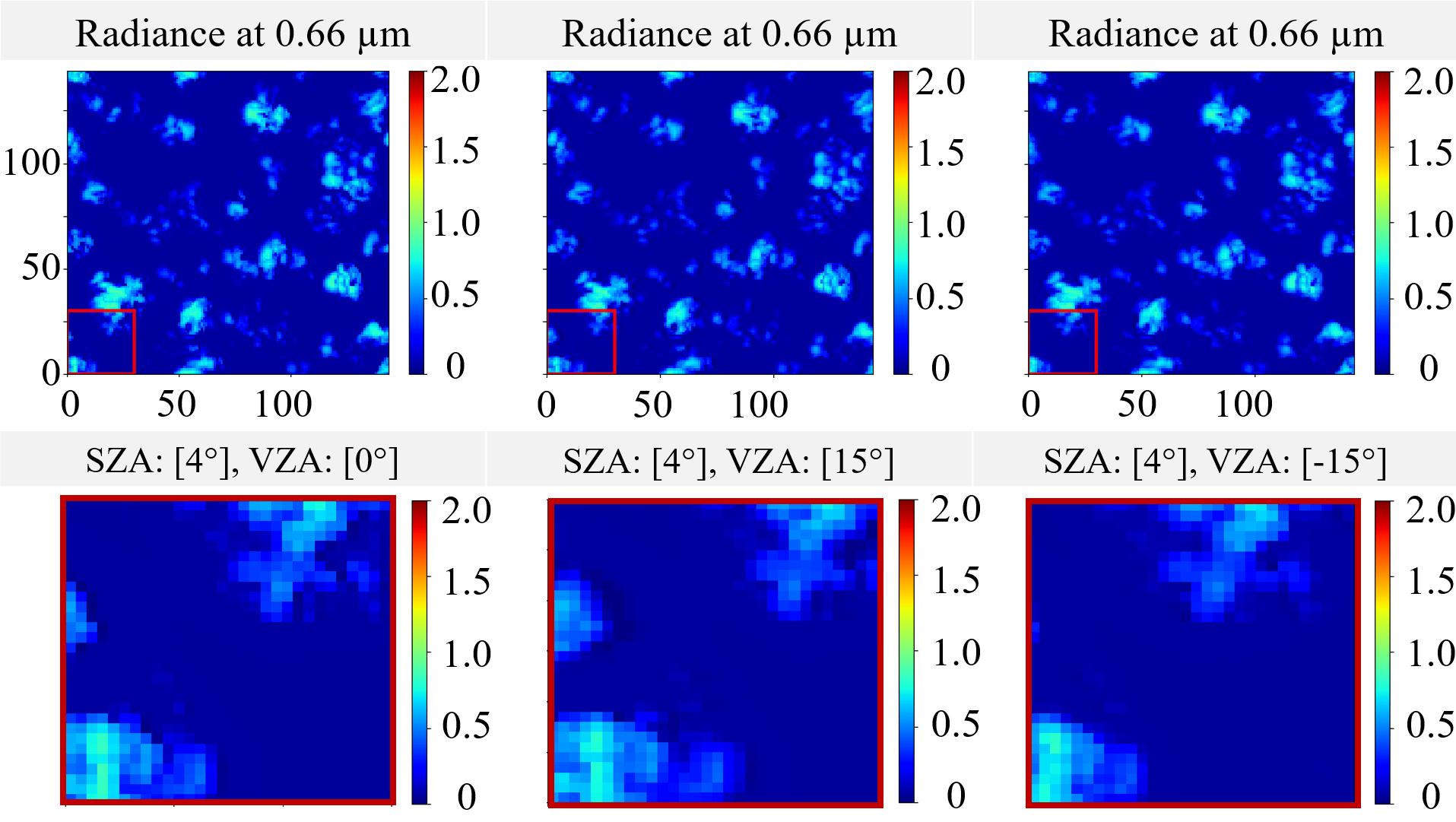}
        \caption{Effects of View Zenith Angle (VZA) on satellite radiance observations. 
        Top row: Radiance observations for the \textit{same cloud field} at different VZAs 
        $[15^\circ, 0^\circ, -15^\circ]$ and SZA $4^\circ$. 
        Bottom row: \textit{Highlighted regions} show distortion and shifting effects at large VZAs.}
        \label{fig:vza_effects}
    \end{subfigure}
    \caption{Comparison of radiance variations due to changes in SZA and VZA for the same cloud field.}
    \label{fig:sza_vza_effects}
    \vspace{-10pt}
\end{figure*}

In this section, we describe our dataset and the challenges associated with cloud property retrievals due to solar and view zenith angles, as well as our problem formulation.

\textbf{Dataset:} Satellites capture cloud radiance observations from real clouds at specific solar zenith angles (SZAs) and view zenith angles (VZAs). However, acquiring radiance measurements from multiple real clouds at varying SZAs and VZAs simultaneously is currently infeasible, as it would require identical satellites positioned at multiple angles.
To systematically evaluate the performance of cloud optical thickness (COT) retrieval algorithms across multiple angles, we generated a real-world cloud-inspired synthetic dataset by simulating radiance observations for each cloud profile at various angles. 
For this study, 102 cloud fields were modeled by using the cloud Liquid Water Content (LWC) form, the Large-Eddy Simulation (LES) ARM Symbiotic Simulation and Observation (LASSO)~\cite{gustafson2020large}, and a constant cloud droplet effective radius of 12$\mu$m for each column across all profiles. The LES profiles from LASSO simulations provide a realistic representation of the atmosphere and clouds, aligning with observations from the Atmospheric Radiation Measurement (ARM) program. The LASSO LES cloud field features a large domain size of $14$km $\times 14 $km$ \times 15 $km, with a horizontal resolution of $100m$ and a vertical resolution of $30m$ below $5km$, extending to $300m$ above $5km$. The combination of a large domain size and high resolution makes LASSO LES cloud fields well-suited for three-dimensional (3D) effects radiative transfer studies. We performed 3D radiance computations at \(0.66 \, \mu m\) and \(2.13 \, \mu m\) using the SHDOM method developed by Evans et al.~\cite{evans1998spherical}. The angular resolution of the SHDOM used was N$\mu$=12  and N=24. The radiative transfer calculation was performed at solar zenith angles (SZAs) $60^\circ, 40^\circ, 20^\circ$, and $4^\circ$, solar azimuth angle (SAA) $0^\circ$ and $180^\circ$, and view zenith angle (VZAs) $15^\circ$ and $0^\circ$ degrees with double periodic horizontal boundary conditions. Note that to indicate profiles with SAA $180^\circ$, we placed a minus ($-$) in front of VZAs. Hence there are three VZAs $[15^\circ, 0^\circ, -15^\circ]$ in total. For each cloud profile there are $4 \times 3 = 12$ radiance observations. The surface was treated as a Lambertian surface with surface albedo of $0.05$.

\textbf{Effects of Solar Zenith Angle: } In Fig.~\ref{fig:sza_effects}, radiance observations at wavelength $0.66\mu m$ is shown for three SZAs $[4^\circ, 40^\circ, 60^\circ]$ at nadir VZA $[0^\circ]$. It is noticeable that the intensity of the radiances change across different SZAs even though the underlying cloud properties are the same. Also there are shadows around the cloudy regions. In order to retrieve the cloud properties accurately, the retrieval method needs to account for this effect of solar zenith angle.

\textbf{Effects of View Zenith Angle: } The radiance observations experience more challenging effects as the viewing angle gets larger and away from the nadir view $[0^\circ]$. An example is shown in Fig.~\ref{fig:vza_effects} where the top row shows the radiance observations for the same cloud profile at different view zenith angles $0^\circ, 15^\circ$ and $-15^\circ$. It is noticeable that the clouds appears to be shifted to the right as the angle increases and vice versa when the angles are negative. 
To accurately retrieve the cloud property COT, the retrieval methods need to account for these apparent deformations in the clouds. 

\textbf{Problem Formulation} We formulate the COT retrieval problem as a regression task where the input to the model is the radiance data from two wavelengths and the output is the COT values.

\section{Cloud-Attention-Net with Angle Coding (CAAC)}
\begin{figure*}[ht]
\centerline{\includegraphics[width=14cm]{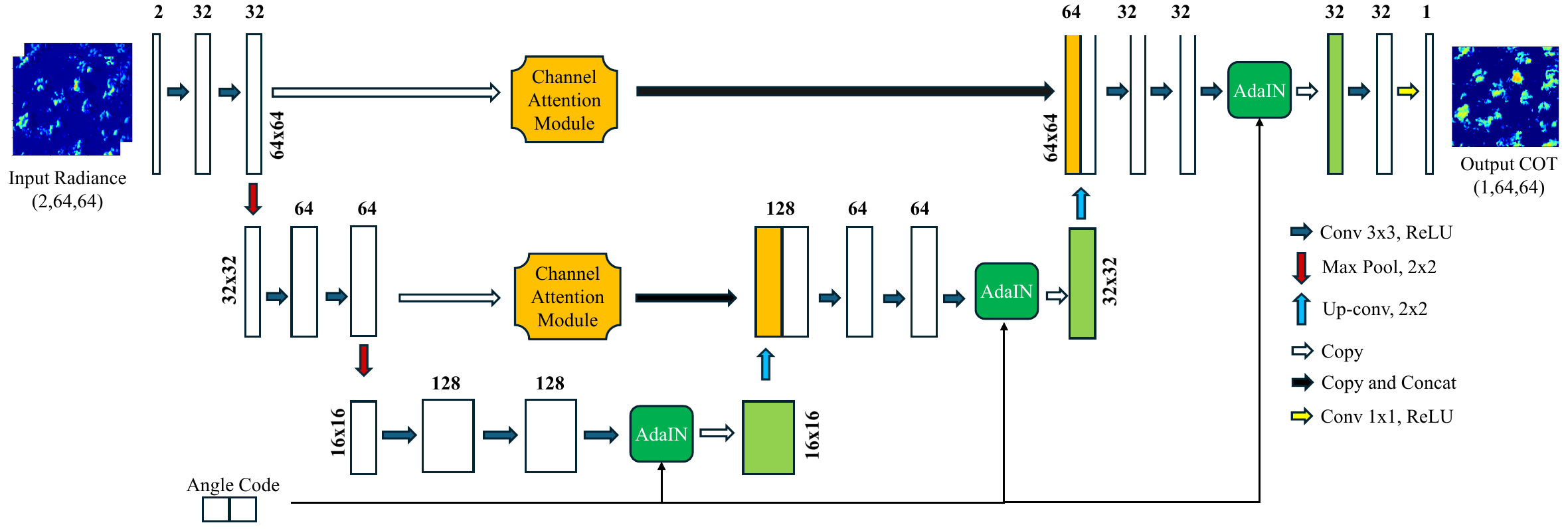}}
\caption{Our Proposed Cloud-Attention-Net with Angle Coding (CAAC) model.}
\label{fig:caac}
\vspace{-12pt}
\end{figure*}

Recent advancements in cloud property retrieval algorithms from radiance observations have increasingly leveraged machine learning and deep learning techniques.
Among these, UNet-style architectures such as CloudUNet~\cite{tushar2024cloudunet} have gained prominence due to their ability to extract 2D spatial features from radiance observations, effectively mitigating 3D radiative transfer effects. To further enhance retrieval accuracy and address the influence of solar and view zenith angles, we propose a novel angle-invariant \textbf{Cloud-Attention-Net with Angle Coding (CAAC) }model, which integrates an attention mechanism and angle coding to improve feature extraction and robustness. The CAAC architecture, illustrated in Fig.~\ref{fig:caac}, introduces two key modifications over standard UNet model:
(1) \textbf{Channel Attention Mechanism}: Instead of traditional skip connections, we incorporate a \textbf{Channel Attention Module (CAM)}~\cite{woo2018cbam} to refine the features extracted at earlier layers. The CAM dynamically reweights feature maps across channels, emphasizing important features while suppressing less relevant ones. This selective refinement enhances feature quality and ensures more informative representations are propagated to deeper layers.
(2) \textbf{Angle Coding for View Geometry Adaptation}: To address variations in radiance observations due to changes in SZA and VZA, we introduce angle coding as an additional input. By explicitly encoding angular dependencies, CAAC learns to adjust feature representations based on the observation geometry, improving robustness across different viewing conditions.
By integrating these components, CAAC effectively enhances cloud optical thickness (COT) retrieval, reducing errors introduced by 3D cloud structures and geometric constraints in satellite observations.

We adopted the adaptive instance normalization (AdaIN) technique to incorporate the angle information into our model~\cite{huang2017arbitrary}. Typically, for image-to-image translation tasks, style information is applied to the input image using the AdaIN block~\cite{liu2021multiple,guo20213s,fan2024domain}. This works by normalizing the feature maps of the input image using the mean and standard deviation learned from the style information as shown in Eq.~\ref{eq1}. 
\begin{equation}\label{eq1}
    AdaIN(x,y)=\sigma(y)(\frac{x-\mu(x)}{\sigma(x)})+\mu(y)
\end{equation}
Here, the normalized feature maps (x) are scaled with standard deviation and shifted with mean learned from style information (y). In our case, AdaIN learns the mean and standard deviation using the angle information. This helps to adjust the feature maps to reduce the effects of satellite's viewing geometry.

\section{Experiments}
We conducted a comprehensive set of experiments to address the following research questions: (1) \textit{Model Comparison}: How does our proposed CAAC model perform compared to existing COT retrieval methods?
(2) \textit{Impact of SZA and VZA}: What is the impact of SZA (Solar Zenith Angle) and VZA (View Zenith Angle) on the performance of COT retrieval models?
(3) \textit{Multi-Angle Training}: Does employing a multi-angle training dataset improve COT retrieval under varying SZA and VZA conditions?

\textbf{Data Partition: } 
To study the effects of solar zenith angle and view zenith angle on the radiance observations and the retrieved cloud properties, we conducted separate experiments with three data partition [see Table.~\ref{tab:data_partition}]. In the first setup, we conducted experiments on radiance observations data from four SZAs $[4^\circ, 20^\circ, 40^\circ, 60^\circ]$ and three VZAs $[15^\circ, 0^\circ, -15^\circ]$, giving us overall performance. In the next setup, all four SZAs are used and the VZA is fixed at $0^\circ$. In the last setup, we used radiance observations from three VZAs $[15^\circ, 0^\circ, -15^\circ]$ and fixed SZA $4^\circ$. This is helpful to understand how much SZA and VZA impact the retrieved properties. We restricted our experiments to $15^\circ$ to $-15^\circ$, as larger values of VZAs significantly distort the cloud structure.

\begin{table}[t]
\centering
\caption{\small{Data Parititon}}
\label{tab:data_partition}
\resizebox{0.6\columnwidth}{!}{%
\centering
\begin{tabular}{cc}
\hline
\toprule
\textbf{Name} & 
\textbf{Viewing Geometry} 
\\ \hline
\begin{tabular}[c]{@{}c@{}}SZA: all \\ VZA: all\end{tabular}      &\begin{tabular}[c]{@{}c@{}}SZA: $[4^\circ, 20^\circ, 40^\circ, 60^\circ]$ \\ VZA: $[15^\circ, 0^\circ, -15^\circ]$\end{tabular}                                                                      \\ \hline
\begin{tabular}[c]{@{}c@{}}SZA: all \\ VZA: fixed\end{tabular}      &\begin{tabular}[c]{@{}c@{}}SZA: $[4^\circ, 20^\circ, 40^\circ, 60^\circ]$ \\ VZA: [$0^\circ$]\end{tabular}                                                                      \\ 
\hline
\begin{tabular}[c]{@{}c@{}}SZA: fixed \\ VZA: all\end{tabular}      &\begin{tabular}[c]{@{}c@{}}SZA: [$4^\circ$] \\ VZA: $[15^\circ, 0^\circ, -15^\circ]$\end{tabular}                                                                      \\ 
\hline
\end{tabular}%
}
\vspace{-20pt}
\end{table}

\textbf{Data Pre-processing: }
The cloud optical thickness (COT) have a range between $[0, 300]$ and have a tail distribution with most of the values residing in a small range $[0,50]$. Therefore we used a shifted log transformation to achieve an evenly distributed representation of the data. Mathematically, $\textit{transformed COT} = log(COT+1)$

\textbf{Implementation details: }
All the models are trained using L2 loss with an early stopping criteria for 50 epochs patience. The learning rate is searched in $[1e^-3, 1e^-2, 1e^-1]$ with optional learning rate scheduler ReduceLR. The batch size is set to $256$, and the training was optimized using Adam optimizer with default parameters. All models were trained on single 24GB RTX6000 and single 48GB RTX8000 GPUs.
We have employed a window based retrieval framework which is a common practice for deep learning based COT retrievals~\cite{okamura2017feasibility, nataraja2022segmentation, tushar2024cloudunet}. We have adopted the framework outlined in ~\cite{tushar2024cloudunet} and chosen $[64\times 64]$ windows with a stride of $10$. 

\textbf{Single and Multiple Angle Training Setup:} Single-angle models are trained using radiance observations from a single combination of SZA[$4^\circ$] and VZA[$0^\circ$] while the multi-angle models are trained using radiance observations from multiple SZAs and VZAs as specified in Table.~\ref{tab:data_partition}.

\textbf{Model Comparisons :}
To evaluate the advantages of the proposed CAAC model, we have compared with it the state-of-the-art COT retrieval methods: IPA~\cite{nakajima1990determination}, UNet~\cite{nataraja2022segmentation, zhao2023cloud,li2023transfer}, and CloudUNet~\cite{tushar2024cloudunet}. We used Mean Squared Error (MSE) as the evaluation metrics.

\section{Results and Discussion}

\vspace{-8pt}
Table ~\ref{tab:szavza} shows the model performance (MSE scores) for all the COT retrieval methods across different data partitions. The results indicate that multi-angle training enhances the performance of all COT retrieval methods compared to single-angle training.
Furthermore, our multi-angle CAAC model achieves the best overall performance, demonstrating \textbf{9x, 7x, and 11x} lower MSE than single-angle methods. Additionally, it surpasses other multi-angle trained COT retrieval methods, as CAAC effectively leverages the attention mechanism and angle information to improve accuracy.

\begin{table}[h]
\centering
\caption{Comparison of COT retrieval methods under different SZAs and VZAs (MSE reported).}
\label{tab:szavza}
\resizebox{0.99\columnwidth}{!}{%
\begin{threeparttable}
\begin{tabular}{lcccc}
\toprule
\textbf{Retrieval Methods} & \textbf{Training Conditions} & \textbf{SZA: all, VZA: all} & \textbf{SZA: all, VZA: fixed} & \textbf{SZA: fixed, VZA: all} \\
\midrule
IPA Retrieval~\cite{nakajima1990determination} & - & \(0.4212 \pm 0.3787\) & \(0.1018 \pm 0.0742\) & \(0.4185 \pm 0.3569\) \\
\midrule
UNet~\cite{nataraja2022segmentation, zhao2023cloud,li2023transfer} & Single Angle & \(0.4097 \pm 0.3734\) & \(0.0722 \pm 0.0222\) & \(0.4325 \pm 0.4282\) \\
CloudUNet~\cite{tushar2024cloudunet} & Single Angle & \(0.4097 \pm 0.3727\) & \(0.0778 \pm 0.0265\) & \(0.4259 \pm 0.4205\) \\
\midrule
UNet~\cite{nataraja2022segmentation, zhao2023cloud,li2023transfer} & Multi-Angle & \(0.0678 \pm 0.0473\) & \(0.0126 \pm 0.0073\) & \(0.0482 \pm 0.0372\) \\
CloudUNet~\cite{tushar2024cloudunet} & Multi-Angle & \(0.0778 \pm 0.0444\) & \(0.0125 \pm 0.0067\) & \(0.0467 \pm 0.0358\) \\
\midrule
CAAC$^\alpha$ & Multi-Angle & \(0.0521 \pm 0.0425\) & \(0.0109 \pm 0.0073\) & \(0.0387 \pm 0.0325\) \\
CAAC$^\beta$ & Multi-Angle & \(0.0466 \pm 0.0400\) & \(0.0109 \pm 0.0073\) & \(0.0380 \pm 0.0296\) \\
\textbf{CAAC (Ours)} & \textbf{Multi-Angle} & \(\mathbf{0.0459 \pm 0.0395}\) & \(\mathbf{0.0106 \pm 0.0069}\) & \(\mathbf{0.0358 \pm 0.0295}\) \\
\bottomrule
\end{tabular}
\begin{tablenotes}
\item $\alpha$: With attention, no angle coding.
\item $\beta$: No attention, with angle coding.
\end{tablenotes}
\end{threeparttable}
}
\vspace{-10pt}
\end{table}

\begin{figure}[ht]
    \centering \includegraphics[width=0.99\linewidth]{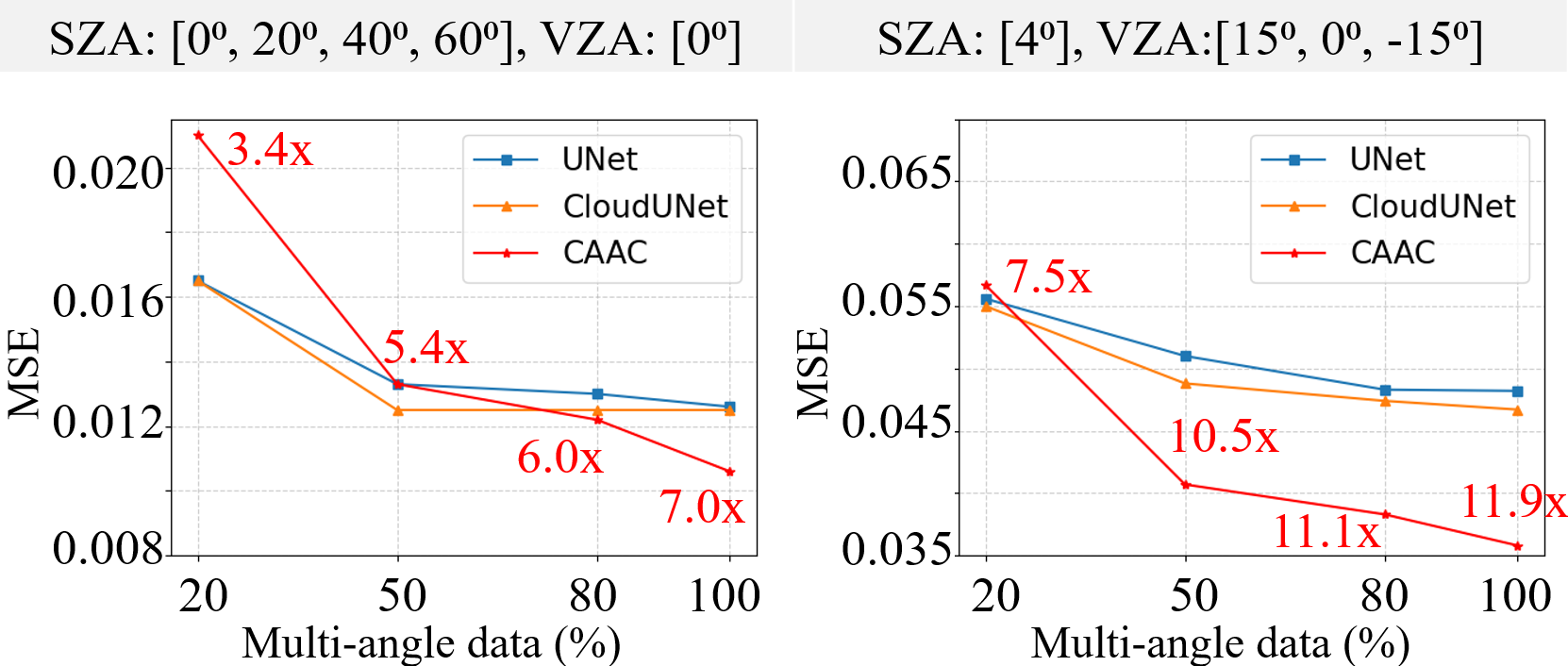}
    \caption{\small{Impact of Multi-angle training.}}
    \label{fig:progress}
    \vspace{-10pt}
\end{figure}

\textbf{Impact of Multi-angle Training}: 
In two data partitions [Fig.~\ref{fig:progress}], we demonstrated that with just 20\% of the multi-angle training data, all models achieve a performance improvement of at least 3.4× and 7.5×, respectively, compared to single-angle deep learning models. As the proportion of multi-angle data increases, our CAAC model outperforms the others, achieving 7× and 11.9× improvement.

\begin{figure}[ht]
    \centering
    \includegraphics[width=0.90\linewidth]{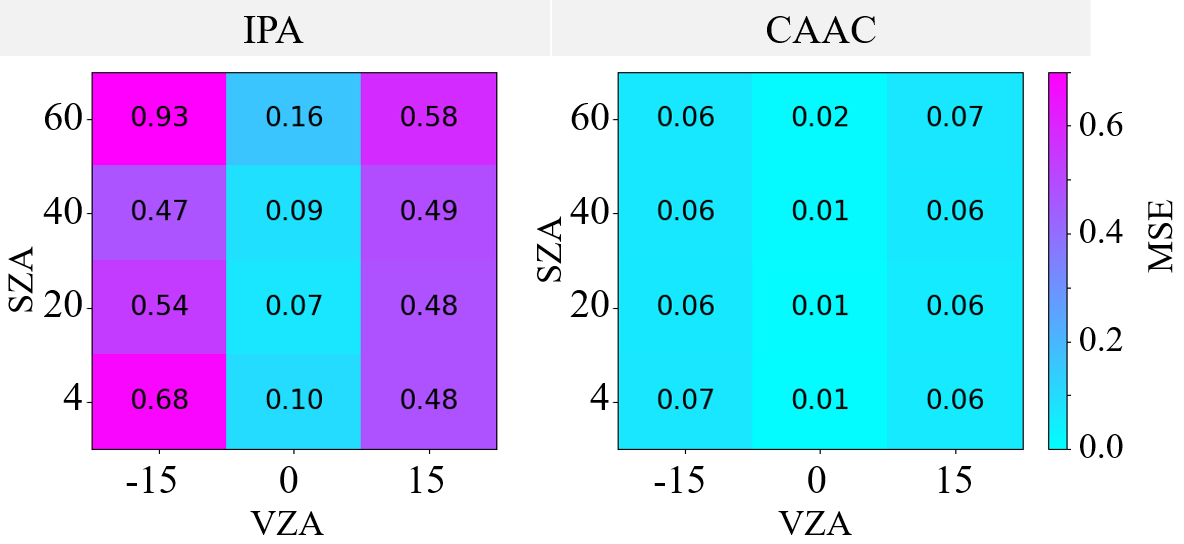}
    \caption{\small{Evaluation of IPA and CAAC on all SZAs and VZAs.}}
    \label{fig:loss_confusion_matrix}
    \vspace{-10pt}
\end{figure}

\textbf{Performance on individual angles: }
Figure~\ref{fig:loss_confusion_matrix} presents the MSE for individual angles across all SZA and VZA for both the IPA and CAAC COT retrieval methods. As expected, COT retrieval becomes increasingly challenging at larger SZAs and VZAs, which is reflected in the higher MSE scores of IPA. In contrast, CAAC effectively mitigates these challenges, achieving significantly lower MSE errors, demonstrating its robustness in varying angular conditions.

\begin{figure}[ht]
    \centering
    \includegraphics[width=0.8\linewidth]{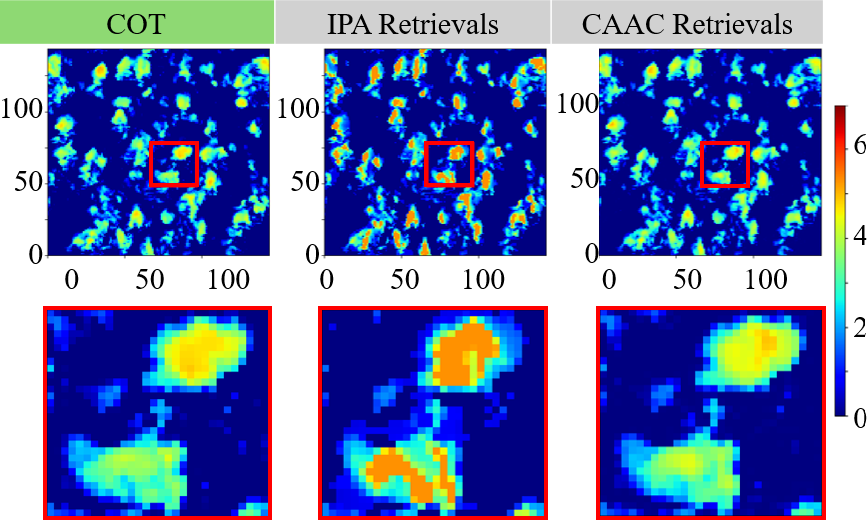}
    \caption{\small{Comparison of COT retrievals methods. First row shows true COT for a cloud profile, COT retrieved by IPA and CAAC (ours) respectively from radiance at SZA $60^\circ$ and VZA $0^\circ$. Bottom row shows the \textit{highlighted region}.}}
    \label{fig:sza_results}
    \vspace{-10pt}
\end{figure}

\begin{figure}[ht]
    \centering
    \includegraphics[width=0.99\linewidth]{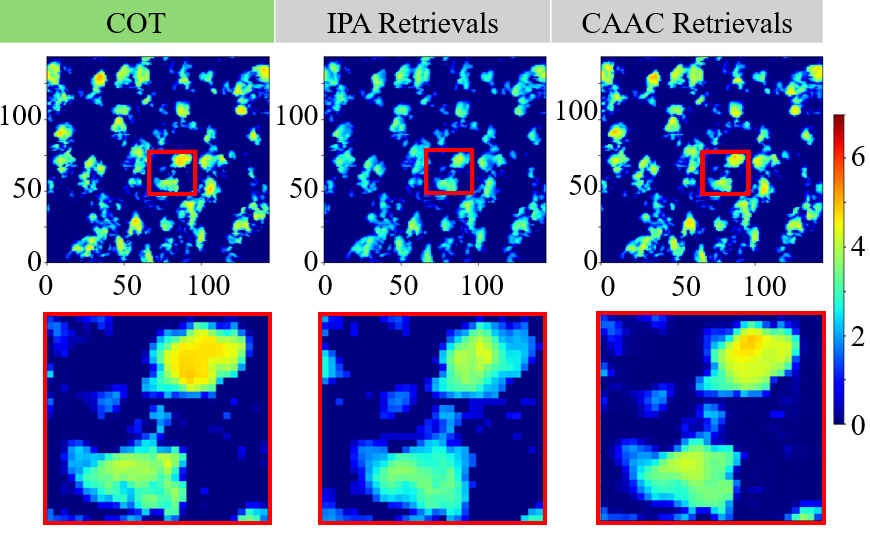}
    \caption{\small{Comparison of COT retrievals methods. First row shows true COT for a cloud profile, Retrieved COT by IPA and CAAC (ours) respectively from radiance at SZA $4^\circ$ and VZA $15^\circ$. Bottom row shows the \textit{highlighted region}.}}
    \label{fig:vza_results}
    \vspace{-10pt}
\end{figure}

\textbf{Qualitative Comparisons: }
Figures~\ref{fig:sza_results} and ~\ref{fig:vza_results} present the retrieved COTs using IPA and our CAAC method, along with the corresponding true COTs for the input radiance observations at [SZA $60^\circ$, VZA $0^\circ$], and [SZA $4^\circ$, VZA $15^\circ$]. 
The satellite’s viewing geometry significantly impacts radiance intensity, cloud structure distortion, and cloud position shifts. In the highlighted region of Fig.~\ref{fig:sza_results}, at larger SZA $60^\circ$, IPA exhibits both overestimation and underestimation in different parts of the top cloud. 
This occurs because a large SZA alters radiance intensity, making some regions appear brighter or darker, leading IPA to misinterpret these areas and incorrectly retrieve COT values. Similarly, as VZA increases, clouds become shifted and deformed, further affecting COT retrieval. In Fig.~\ref{fig:vza_results}, this deformation introduces errors in IPA-retrieved COTs, particularly on the left side, where erroneous COT values appear in areas where they should not be.
In contrast, our CAAC model effectively mitigates these challenges, demonstrating greater robustness by accounting for the effects of varying SZA and VZA, leading to more accurate COT retrievals compared to IPA.

\begin{figure}[ht]
    \centering
    \includegraphics[width=0.99\linewidth]{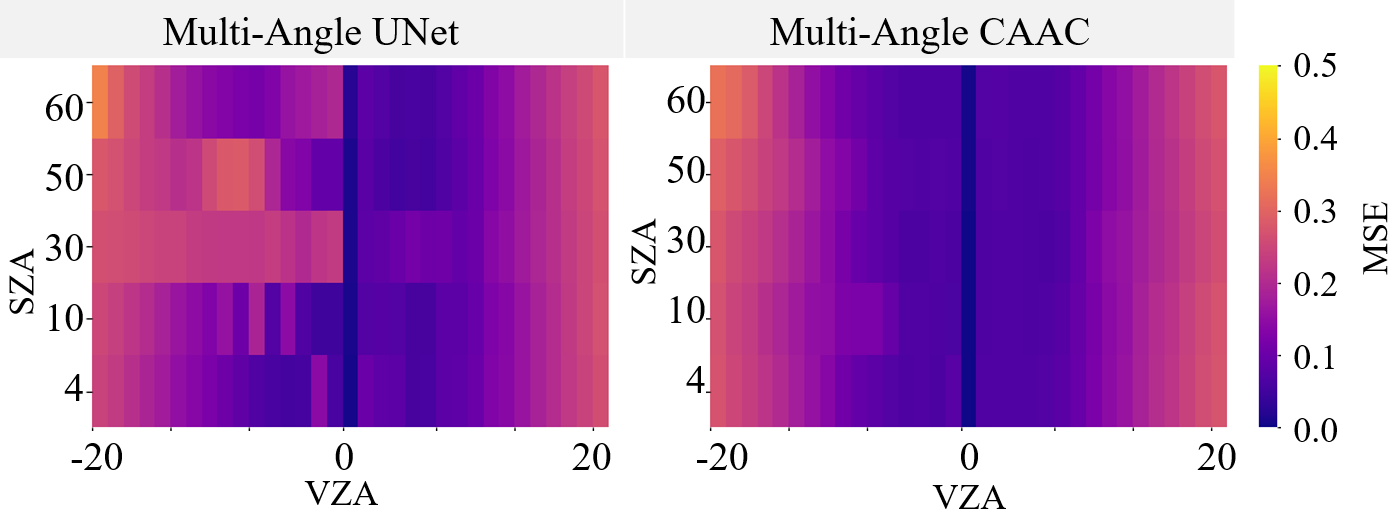}
    \caption{Performance evaluation on radiance observations from five SZAs $[5,10,20,30,50]$ and thirty three VZAs $[20,19,18,...,7,6,5,0,-5,-6,-7,...,-18,-19,-20]$.}
    \label{fig:overall}
    \vspace{-15pt}
\end{figure}

\textbf{Evaluation on unseen SZA and VZA: } 
We have established that the multi-angle training is helpful to tackle the effects of solar and view zenith angles. But the models were tested on the same set of angles as they were trained on. \textit{Can the model reduce these effects for angles that were not present in the training set which is likely to be the case in real-world?} To answer this question, we have trained UNet and CAAC using multi-angle training and evaluated them on a different set of LES cloud profiles with radiance observations at five SZAs $[5^\circ, 10^\circ, 20^\circ, 30^\circ, 50^\circ]$ and 
thirty three VZAs $[20^\circ, 19^\circ, 18^\circ, 
..., 6^\circ, 5^\circ, 0^\circ, -5^\circ, -6^\circ,$ 
$... , -18^\circ, -19^\circ, -20^\circ]$.

Given the large number of angle combinations [$total =5\times 33= 165$], we present the results as a heatmap in Fig.~\ref{fig:overall}. Our multi-angle CAAC model achieved the best and most stable performance, outperforming state-of-the-art UNet methods. Notably, all models performed well at the nadir angle, whereas performance degraded at larger SZAs and VZAs due to their stronger impact on radiance observations.

\vspace{-10pt}
\section{Conclusion}

In this work, we introduced Cloud-Attention-Net with Angle Coding (CAAC)—the first angle-invariant, attention-based deep learning model for COT retrieval. We showed that our CAAC model which comprises of a compact U-Net architecture with attention and angle-coding modules, can effectively leverage angle information from radiance observations to mitigate the impacts of SZA, VZA, and 3D radiative transfer effects, resulting in accurate COT retrievals. Through extensive experiments, we demonstrated that CAAC achieves state-of-the-art performance under varying SZA, VZA, and multi-angle training conditions. Notably, our model delivers at least a nine-fold improvement (in terms of MSE) compared to both physics-based IPA retrieval and state-of-the-art deep learning COT retrieval methods. In future work, we plan to extend CAAC for jointly retrieval of COT with cloud effective radius in multi-angle scenarios.

\vspace{-10pt}
\section*{Acknowledgments}
This research is partially supported by grants from NSF 2238743 and NASA 80NSSC21M0027.

This work was carried out using the computational facilities of the High Performance Computing Facility, University of Maryland Baltimore County. - https://hpcf.umbc.edu/

\small
\bibliographystyle{IEEE}
\bibliography{ref}

\end{document}